\documentclass[runningheads]{llncs}
\usepackage[T1]{fontenc}
\usepackage{graphicx}

\usepackage{algorithm}
\usepackage{algorithmic}

\usepackage{marvosym}
\usepackage{array}
\usepackage{booktabs}
\usepackage{amsfonts}
\usepackage{amsmath}
\usepackage{amssymb}
\usepackage{multirow}
\usepackage{tabularx}
\usepackage{makecell}
\usepackage{xcolor}
\usepackage{colortbl,booktabs}

\begin{document}
\title{MCN-CL: Multimodal Cross-Attention Network and Contrastive Learning for Multimodal Emotion Recognition}
\titlerunning{MCN-CL for Multimodal Emotion Recognition}
%

\author{Feng Li\inst{1} \and
Ke Wu\inst{1} \and
Yongwei Li\inst{2}\textsuperscript{(\Letter)}}
\authorrunning{F. Li et al.}
%
\institute{School of Computer and Information Engineering, Anhui University of Finance and Economics, Anhui, 233030, China
\\
 \and
State Key Laboratory of Cognitive Science and Mental Health, Institute of Psychology, Chinese Academy of Sciences, Beijing, 100045, China\\
\email{liyw}@psych.ac.cn}
\maketitle              
\begin{abstract}
Multimodal emotion recognition plays a key role in many domains, including mental health monitoring, educational interaction, and human-computer interaction. However, existing methods often face three major challenges: unbalanced category distribution, the complexity of dynamic facial action unit time modeling, and the difficulty of feature fusion due to modal heterogeneity. With the explosive growth of multimodal data in social media scenarios, the need for building an efficient cross-modal fusion framework for emotion recognition is becoming increasingly urgent. To this end, this paper proposes Multimodal Cross-Attention Network and Contrastive Learning (MCN-CL) for multimodal emotion recognition. It uses a triple query mechanism and hard negative mining strategy to remove feature redundancy while preserving important emotional cues, effectively addressing the issues of modal heterogeneity and category imbalance. Experiment results on the IEMOCAP and MELD datasets show that our proposed method outperforms state-of-the-art approaches, with Weighted F1 scores improving by 3.42\% and 5.73\%, respectively.

\end{abstract}
%
%
%


\section{Introduction}

Emotion recognition technology, as a cutting-edge field in human-computer interaction research, is evolving from unimodal analysis to multimodal fusion \cite{li2025mimcl,Erkantarci2023AnES,li2025wavfusion}. With the widespread adoption of mobile internet technology, video calls and voice messages have become the mainstream forms of interaction on social media \cite{Park2016MultimodalAA}, dominating modern communication methods. Additionally, psychological research findings indicate that during emotional expression, some users primarily rely on paralinguistic cues, such as the melody, tone, and rhythm of speech, to convey emotions, while others tend to express inner feelings more through facial micro-expressions \cite{Le2010ASO}. These converging trends underscore the growing urgency to develop sophisticated multimodal emotion recognition systems. By integrating complementary data streams, such systems aim to overcome the inherent limitations of unimodal analysis, thereby capturing the complex, interwoven nature of human emotions with greater accuracy and depth for a more holistic and reliable understanding of affective states.

Although significant progress has been made in multimodal emotion recognition, current cross-modal fusion methods still face three fundamental limitations that hinder their effective promotion in practical applications. First, the redundancy and heterogeneity problems in cross-modal fusion need urgent resolution. Traditional multimodal fusion methods, such as the MVN-based feature connection \cite{Ma2021AMN} or the EmoCaps capsule network \cite{Li2022EmoCapsEC}, fail to model the differences between modalities. This results in a masking effect where large amounts of redundant cross-modal information cover up key affective patterns \cite{Cambria2017BenchmarkingMS,Wang2023MultimodalSA}. For example, Hu et al. \cite{Hu2022MMDFNMD} developed MM-DFN, which employs gating mechanisms to dynamically adjust modal weights. However, its feature connection strategy still cannot eliminate the spectral semantic coupling noise between text and speech \cite{Ghosh2021DepressionIE}. Second, there is a clear deficiency in modeling dynamic micro-expressions. Current visual architectures, such as the ResNet-101 model \cite{He2015DeepRL}, primarily rely on global pooling or static attention mechanisms. This makes it difficult for them to accurately capture the spatiotemporal dynamic features of facial action units (FAUs), thus leading to inaccurate recognition and analysis of micro-expressions. Finally, the discriminative bias problem is caused by extreme category imbalance. Cross-entropy loss-based methods, such as MultiEMO \cite{Shi2023MultiEMOAA} and TelME \cite{Yun2024TelMETM}, yield low F1 scores for sparse classes.

Therefore, to tackle multimodal emotion recognition challenges, this paper proposes the Multimodal Cross-Attention Network and Contrastive Learning (MCN-CL). Our model utilizes cross-attention to effectively integrate features from acoustic, visual, and linguistic modalities, while contrastive learning refines the latent representations to better distinguish between emotional states. The main contributions are summarized as follows

\begin{itemize}
	\item We propose a PSA-enhanced Visual Feature Extraction Module. This module precisely captures facial dynamic patterns and enhances the resolution of emotional features through multi-scale feature extraction and channel attention interaction.
	\item We present a multimodal cross-attention network and contrastive learning enhancement module, which progressively integrates information from text, audio, and visual modalities via multi-layer bidirectional multi-head cross-attention mechanisms to generate comprehensive feature representations. 
	\item We introduce a contrastive learning enhancement module. By integrating a hard negative sample mining strategy, it strengthens the model's capacity to distinguish difficult samples and refines the alignment of cross-modal representations.
	\item We conduct extensive experiments on the IEMOCAP and MELD datasets, respectively. The results demonstrate that our approach outperforms previous state-of-the-art methods. 
\end{itemize}

\section{Proposed Method}

This section details the method proposed in this paper, whose core architecture is shown in Fig. 1. It mainly consists of three steps: unimodal feature extraction, multimodal cross-attention network, and contrastive learning enhancement.

\begin{figure*}[t]
	\centering
	\includegraphics[width=1.1\textwidth]{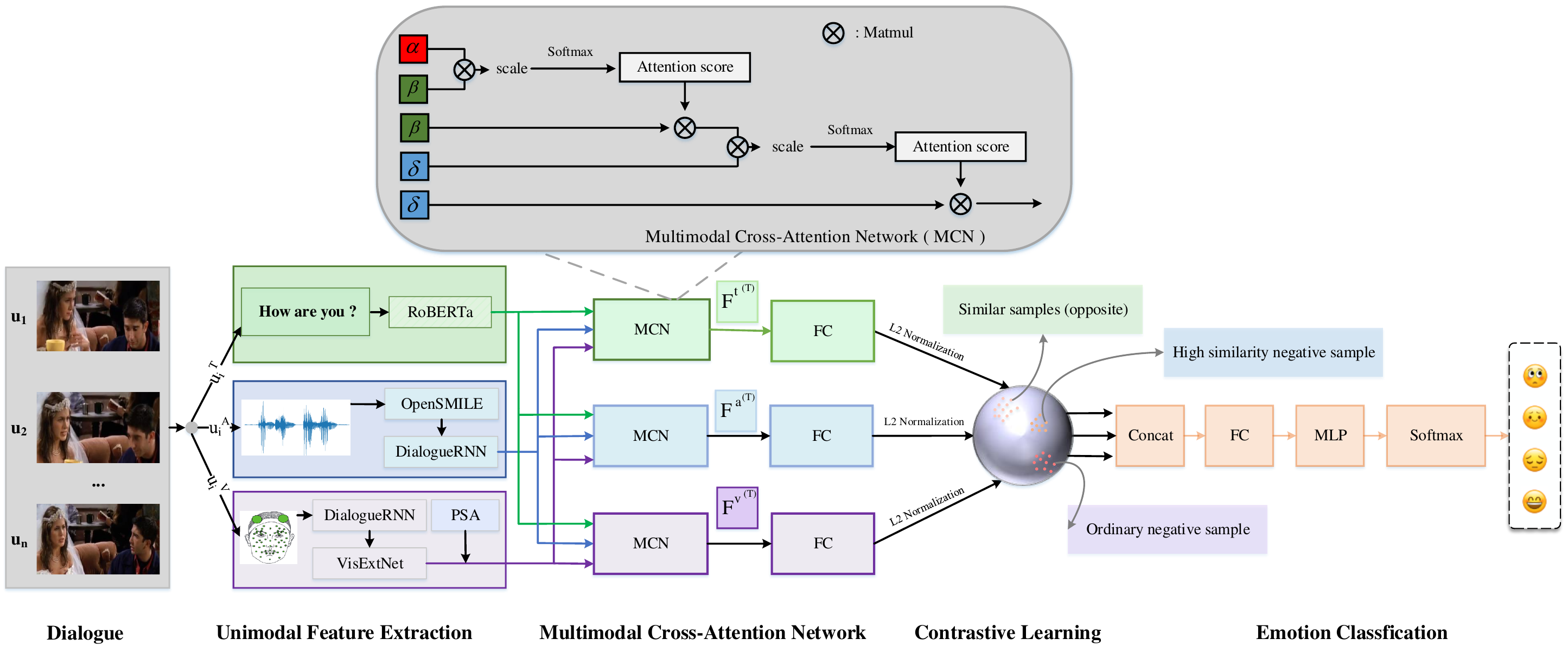}
	\caption{The framework of our proposed method.}
	\label{fig:speech_production}
\end{figure*}

\subsection{Task Definition}

Given a dialogue sequence containing N speakers, denoted as $U=\left\{u_1, u_2, \ldots, u_n\right\}$, each utterance $u_i$ consists of three modalities: visual$\left(u_i^V\right)$, speech$\left(u_i^A\right)$, and text $\left(u_i^T\right)$. The objective is to predict the corresponding emotional category for each utterance $y_i \in$ $\{1,2, \ldots, K\}$, and $K$ is the total number of emotion categories.

\begin{figure*}[t]
	\centering
	\includegraphics[width=1.05\textwidth]{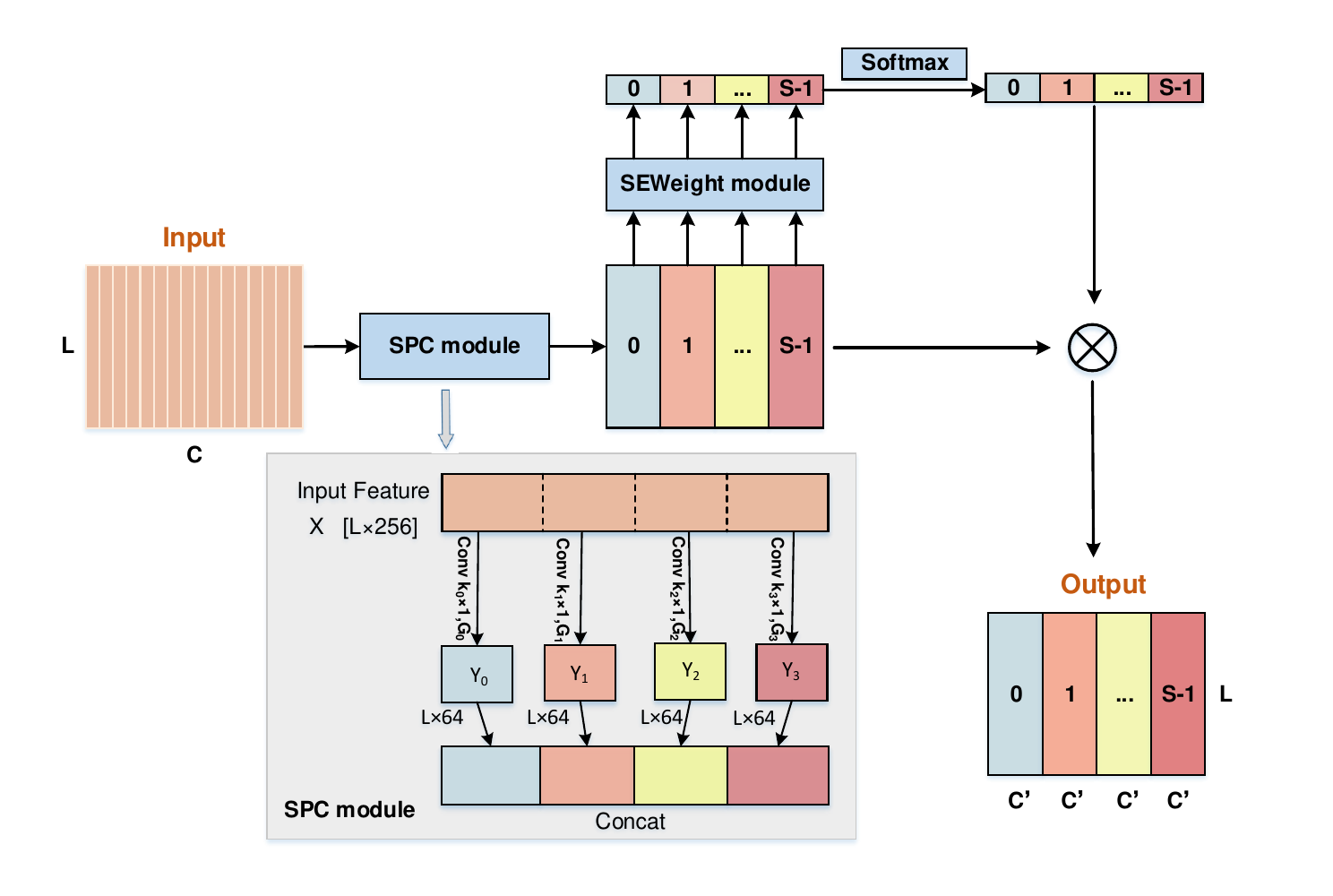}
	\caption{The overview of PSA-enhanced Visual Feature Extraction Module.}
	\label{fig:speech_production}
\end{figure*}

\subsection{PSA-enhanced Visual Feature Extraction Module}
In this study, we design a one-dimensional PSA module for time-series visual feature modeling tasks in Fig.2. By leveraging multi-scale spatiotemporal feature decomposition and dynamic channel attention recalibration, this module effectively alleviates the problems of limited receptive field and local detail loss commonly found in traditional models.
\subsubsection{Multi-scale one-dimensional temporal convolution operation}
To extract the dynamic patterns of FAUs in the temporal dimension, this section extends the original PSA's two-dimensional spatial convolution to a one-dimensional temporal convolution. The input feature tensor is denoted as $X \in R^{L \times C}$, where L represents the number of video frames and $\mathrm{C}=256$ indicates the number of feature channels.

We design S=4 parallel sets of standard convolution kernels with different kernel sizes, specifically $3 \times 1,5 \times 1,7 \times 1$, and $9 \times 1$,to extract sequence features ranging from short-term to long-term. Following the grouped convolution strategy, each set processes $\mathrm{C} / \mathrm{S}$ channels. The feature output of the s-th branch is expressed as follows

\begin{eqnarray}
	Y_s=\operatorname{Conv}_s\left(X_s\right)
\end{eqnarray}
where $X_s$ is the input of the s-th branch. $\operatorname{Conv}_s(\cdot)$ represents a standard convolution operation with a kernel size of $2 *(s+1)+1$, and $Y_s$ is the feature output for each scale.

Finally, all scale outputs are concatenated to obtain the fused feature $Y$ as follows
\begin{eqnarray}
	Y=\operatorname{Concat}\left(Y_0, Y_1, Y_2, Y_3\right)
\end{eqnarray}

\subsubsection{Channel Attention Interaction}

To dynamically adjust the importance of each channel, we introduce the SEWeight mechanism \cite{Erkantarci2023AnES} to weight the fused feature Y. First, global average pooling (GAP) is applied to Y along the time dimension to obtain the channel-level representation as follows
\begin{eqnarray}
	\mathrm{Z}=\mathrm{GAP}(\mathrm{Y})
\end{eqnarray}

In this context, it reduces the temporal dimension (L) of $Y$ to 1, resulting in a channel-level representation $Z \in R^{1 \times c}$, where each element $Z_c$ represents the average value of channel c across all time steps. This allows the model to focus on the most relevant channels for the given task.

Subsequently, it $Z$ is passed through two fully connected layers to generate an attention weight vector as follows
\begin{eqnarray}
	\mathrm{W}=\operatorname{Sigmoid}\left(\mathrm{FC}_2\left(\operatorname{ReLU}\left(\mathrm{FC}_1(\mathrm{Z})\right)\right)\right)
\end{eqnarray}
where $\mathrm{FC}_1$ and $\mathrm{FC}_2$ denote the fully connected layer. $\mathrm{FC}_1$ transforms the channel-level representation Z from dimension C to $\mathrm{C} / 4$. The ReLU activation function is applied after $\mathrm{FC}_1$ to introduce non-linearity. $\mathrm{FC}_2$ then transforms the output back to dimension C. The Sigmoid activation function is applied after $\mathrm{FC}_2$ to produce the channel attention coefficient $W$, where each element $W_c$ indicates the importance weight of the channel $c$.

Finally, the attention coefficient W is applied to the original fusion feature Y through element-wise multiplication in broadcast mode as follows
\begin{eqnarray}
	C^v=\mathrm{Y} \odot \mathrm{~W}
\end{eqnarray}
where $\odot$ represents per-channel multiplication. This operation adjusts the importance of each channel $Y$ based on the learned attention weights $W$. And $C^v$ is aligned with the dimensions of the text and speech modalities to facilitate subsequent cross-modal interaction operations.

\subsection{Multimodal Cross-Attention Network}
The three modalities of text, audio, and visual, bidirectional multi-head cross-attention are first used to achieve cross-modal information interaction between any two modalities. Taking the text modality as an example, the features obtained from its interaction with the audio modality are then input into another bidirectional multi-head cross-attention as queries (Q). Here, the visual modality serves as keys (K) and values (V) for further feature fusion and refinement. Each interaction involves different modality combinations, ensuring that each modality can obtain unique context information from the others. This process enables more comprehensive and in-depth cross-modal information fusion.

First, multi-head projection transformations are applied separately to the query, key, and value as follows
\begin{eqnarray}
	Q_h^{ta^{(j)}}=F^{t^{(j-1)}} W^{Q_h^{ta^{(j)}}}
\end{eqnarray}
\begin{eqnarray}
	K_h^{ta^{(j)}}=C^aW^{K_h^{ta^{(j)}}}
\end{eqnarray}
\begin{eqnarray}
	V_h^{t a^{(j)}}=C^aW^{V_h^{t a^{(j)}}}
\end{eqnarray}
where $W^{Q_h^{ta^{(j)}}}, W^{K_h^{ta^{(j)}}}$ and $W^{V_h^{ta^{(j)}}}$ are the parameters of the multi-head projection matrix, H is the number of attention heads, and $h \in\{1, \ldots H\}$.

Then, for each attention head, the weighted attention output is computed as follows
\begin{eqnarray}
	A_h^{t a^{(j)}}=\operatorname{Softmax}\left(\frac{Q_h^{t a^{(j)}} K_h^{t a^{(j)}}}{\sqrt{d_{K_h^{t a^{(j)}}}}}\right) V_h^{t a^{(j)}}, h \in\{1, \ldots H\}
\end{eqnarray}

All attention heads are concatenated as follows
\begin{eqnarray}
	MH^{t a^{(j)}}=\operatorname{Concat}\left(A_1^{t a^{(j)}}, A_2^{t a^{(j)}}, \ldots, A_H^{t a^{(j)}}\right) W^{O^{t a^{(j)}}}
\end{eqnarray}
where $W^{O^{ta^{(j)}}}$ is the output projection matrix.

The updated feature representation is as follows
\begin{eqnarray}
	F^{t a v^{(j)}}=\operatorname{Add} \& \operatorname{Norm}\left(F^{t a^{(j)}}+M H^{t a v^{(j)}}\right)
\end{eqnarray}

Next, perform multi-head attention computation on the updated text modality and visual modality as follows
\begin{eqnarray}
	Q_h^{\operatorname{tav}^{(j)}}=F^{\operatorname{ta}^{(j)}}W^{K_h^{\operatorname{tav}^{(j)}}}
\end{eqnarray}
\begin{eqnarray}
	K_h^{\operatorname{tav}(j)}=C^v W^{Q_h^{\operatorname{tav}(j)}}
\end{eqnarray}
\begin{eqnarray}
	V_h^{\operatorname{tav}(j)}=C^v W^{v_h^{\operatorname{tav}(j)}}
\end{eqnarray}
where $W^{Q_h^{\operatorname{tav}(j)}}, W^{K_h^{\operatorname{tav}(j)}}$ and $W^{v_h^{\operatorname{tav}(j)}}$ are the parameters of the multi-head projection matrix.

By stacking $T$ layers of this mechanism, the MCN can iteratively refine cross-modal representations, leading to more robust and context-aware feature fusion. The value of T is dataset-dependent; details are provided in the ablation study section. Each layer builds upon the outputs of the previous one, enabling the network to capture both local and global dependencies across modalities. This layered architecture not only enhances the model's expressive capacity but also improves its adaptability to complex multimodal data.

Similarly, the audio and visual modalities are also calculated with cross-attention with other modalities, obtaining updated representations $F^{a^{(j)}}$ and $F^{v^{(j)}}$. The multi-head bidirectional design can capture fine-grained inter-modal dependencies and ensures smooth information reconstruction through residual normalization and feedforward networks.

\subsection{Contrastive Learning}

For the text, audio, and visual features extracted by the MCN: $F^{a^{(T)}} F^{t^{(T)}} F^{v^{(T)}}$ we introduce a contrastive learning mechanism to further optimize the consistency of crossmodal representations. Specifically, we perform contrastive learning on the features of each modality separately to enhance the model's understanding and discrimination of the intrinsic structures across modalities. The mapped features are L2-normalized so that they are on the unit sphere, which facilitates the subsequent calculation of cosine similarity as follows
\begin{eqnarray}
	z_{\text {text }}=\operatorname{ReLU}\left(W_{\text {text\_} fc 1} F^{t^{(T)}}+b_{\text {text\_} f c 1}\right) 
\end{eqnarray}
\begin{eqnarray}
	z_{\text {text }}=W_{\text {text\_} f c 2} z_{\text {text }}+b_{\text {text\_} f c 2} 
\end{eqnarray}
\begin{eqnarray}
	z_{\text {audio }}=\operatorname{ReLU}\left(W_{\text {audio\_} f c 1} F^{a^{(T)}}+b_{\text {audio\_} f c 1}\right)
\end{eqnarray}
\begin{eqnarray}
	z_{\text {audio }}=W_{\text {audio\_} f c 2} z_{\text {audio }}+b_{\text {audio\_} f c 2}
\end{eqnarray}
\begin{eqnarray}
	z_{\text {visual }}=\operatorname{ReLU}\left(W_{\text {visual\_} f c 1} F^{v^{(T)}}+b_{\text {visual\_} f c 1}\right)
\end{eqnarray}
\begin{eqnarray}
	z_{\text {visual }}=W_{\text {visual\_} f c 2} z_{\text {visual }}+b_{\text {visual\_} f c 2} 
\end{eqnarray}
where $W_{(\*)_f c 1}$ and $W_{(\*)_f c 2}$ are the weight matrices of the fully connected layer, and $b_{(\*)_f c 1}$, $b_{(\*)_f c 2}$ are the bias terms.

Subsequently, the mapped features are L2-normalized so that they are on the unit sphere, which facilitates the subsequent calculation of cosine similarity as follows
\begin{eqnarray}
	z_{\text {text }}^{\prime}=\frac{z_{\text {text}}}{\left\|z_{\text {text }}\right\|_2}
\end{eqnarray}
\begin{eqnarray}
	z_{\text {audio }}^{\prime}=\frac{z_{\text {audio }}}{\left\|z_{\text {audio}}\right\|_2}
\end{eqnarray}
\begin{eqnarray}
	z_{\text {visual }}^{\prime}=\frac{z_{\text {visual}}}{\left\|z_{\text {visual}}\right\|_2}
\end{eqnarray}

For each modality, we construct a supervised contrastive loss function, designed to bring closer the distances between samples of the same class while pushing farther apart the distances between samples of different classes. For the i-th sample, its corresponding modality's loss function is defined as follows

\begin{eqnarray}
	L_{\text {contrast }}=-\frac{1}{\left|\mathcal{P}_i\right|} \sum_{p \in \mathcal{P}_i} \log \frac{\exp \left(\operatorname{sim}\left(z_i, z_p\right) / \tau\right)}{\sum_{k=1}^N 1_{[k \neq i]} \exp \left(\operatorname{sim}\left(z_i, z_k\right) / \tau\right)+\sum_{p \in \mathcal{P}_i} \exp \left(\operatorname{sim}\left(z_i, z_p\right) / \tau\right)} 
\end{eqnarray}
where $\mathcal{P}_i$ represents a set of positive samples of the same kind as sample $i \cdot \operatorname{sim}\left(z_i, z_j\right)$ represents the cosine similarity of the features of samples $i$ and $j$ in the latent space. $\tau$ is a temperature parameter used to control the sharpness of the similarity distribution. $1_{[k \neq i]}$ is an indicator function, which takes $k \neq i$ when B is true; otherwise, it takes $0 . \mathrm{N}$ the total number of samples in the batch.

To enhance the model's ability, we introduce a hard negative mining strategy. We select the top $30 \%$ most similar negative sample subset to participate in loss computation, while the remaining negative samples are used solely for normalization. Specifically, for each sample $i$, we compute its similarity with all negative samples and then select the k most similar negative samples to participate in loss computation as follows
\begin{eqnarray}
	\mathcal{N}_i^{\text {hard }}=\left\{j \mid j \notin \mathcal{P}_i \cup\{i\}, \operatorname{sim}\left(z_i, z_j\right) \geq \operatorname{top} k\left(\operatorname{sim}\left(z_i, z_{\mathcal{N}_i}\right)\right)\right\}
\end{eqnarray}

\begin{eqnarray}
	L_{\text {contrast }}^{\text {hard }}=-\frac{1}{\left|\mathcal{P}_i\right|} \sum_{p \in \mathcal{P}_i} \log \frac{\exp \left(\operatorname{sim}\left(z_i, z_p\right) / \tau\right)}{\sum_{n \in \mathcal{N}_i^{\text {hard }}} \exp \left(\operatorname{sim}\left(z_i, z_n\right) / \tau\right)+\sum_{p \in \mathcal{P}_i} \exp \left(\operatorname{sim}\left(z_i, z_p\right) / \tau\right)}
\end{eqnarray}
where $\mathcal{N}_i^{\text {hard }}$ represents the set of difficult negative samples for sample $i$.

Finally, we fuse the contrastive loss of text, audio, and visual modalities to obtain the final contrastive loss as follows

\begin{eqnarray}
	L_{\text {total\_contrast }}=\alpha \cdot L_{\text {contrast\_text }}+\beta \cdot L_{\text {contrast\_audio }}+\gamma \cdot L_{\text {contrast\_visual }} 
\end{eqnarray}
where $\alpha, \beta$, and $\gamma$ are weighting coefficients used to balance the contributions of the modal contrast losses.

\subsection{Emotion Classification}

After optimizing the features through the MCN-CL module, the multimodal feature representations $f_{\text {text }}, f_{\text {audio }}$, and $f_{\text {visual }}$ are concatenated to form a comprehensive feature vector. This vector first passes through a fully connected layer, which integrates and transforms the features to map the high-dimensional concatenated features into a feature space suitable for classification. Then, the feature vector is input into a 2-layer Multilayer Perceptron (MLP) with ReLU activation, where the MLP further extracts and fuses features through nonlinear transformations, enhancing the model's expressive capability
and ability to capture complex patterns. Finally, a Softmax layer is used for emotion recognition, selecting the emotion label with the highest probability as the prediction. The process is illustrated as follows
\begin{eqnarray}
	h_{\text {concat}}=\left[f_{\text {text}}, f_{\text {audio}}, f_{\text {visual}}\right]
\end{eqnarray}
\begin{eqnarray}
	h_{f c}=F C\left(h_{\text {concat}}\right)
\end{eqnarray}
\begin{eqnarray}
	h_{\text {mlp }}=M L P\left(h_{f c}\right)
\end{eqnarray}
\begin{eqnarray}
	\hat{y}=\operatorname{Softmax}\left(W_{\text {softmax}} h_{\text {mlp}}+b_{\text {softmax}}\right)
\end{eqnarray}
where $h_{\text {concat }}$ is the concatenated feature vector, $h_{f c}$ is the output of the fully connected layer, $h_{\text {mlp}}$ is the output of the MLP, and $W_{\text {softmax }}$ and $b_{\text {softmax }}$ are the weight matrix and bias term of the Softmax layer, respectively.

\section{Experiment Evaluation}
\subsection{Datasets}
In this study, we evaluated the proposed method on two datasets: IEMOCAP \cite{Busso2008IEMOCAPIE} and MELD \cite{Poria2018MELDAM}, respectively. 

\textbf{IEMOCAP}: This dataset is a database containing approximately 12 hours of English audio-visual data, encompassing video, speech, facial motion capture, and text transcriptions. The database is structured around dyadic conversations, where actors interact through either improvisation or specifically designed scripted scenarios. These scenarios were carefully selected to elicit emotional expressions. The dataset provides six categorical emotion labels (such as anger, happiness, sadness, frustration, neutral, and excitement) as well as dimensional annotations (including valence, activation, and dominance).

\textbf{MELD}: This dataset is a multimodal, multi-speaker dialogue dataset generated from dialogue scenes of the TV show 'Friends'. The dataset contains a total of 13,708 utterances, 1,433 dialogues, and 304 different speakers. Specifically, unlike two-person dialogue datasets like IEMOCAP, MELD's dialogues include three or more speakers. Each utterance in the dialogue is annotated with one of seven emotional category labels, including anger, disgust, fear, joy, neutral, sadness, and surprise.

\subsection{Setting and Metric}
This study was conducted on PyTorch 2.0.0, Python 3.8.0 and CUDA Toolkit 12.1. A unified cross-modal feature processing flow is adopted in the experiment: text features are extracted by the RoBERTa-base model, and the dimensionality is reduced from 768 to 256 by a fully connected layer; speech features are extracted by OpenSMILE tool, and then linearly transformed to 256 dimensions; visual features are processed by ResNet-101 and Pyramid Squeeze Attention module, and 1000-dimensional features are extracted and projected to 256-dimensional space. On the IEMOCAP and MELD datasets, we adopt the Weighted F1 as evaluation metrics \cite{liang2021attention,lv2021progressive}. 

\subsection{Results and Discussion}
In this section, the experiment results are reported to evaluate the proposed method on the IEMOCAP and MELD datasets, as shown in Table 1 and Table 2, respectively.

On the IEMOCAP dataset in Table 1, the model achieved a Weighted F1 score of 74.22\%. Meanwhile, on the MELD dataset, the Weighted F1 score reached as high as 73.1\%, significantly outperforming the second-best method TelME (67.37\%) by 5.73\%. This consistently excellent performance across datasets validates the generalization capability of the framework design, particularly demonstrating advantages in multimodal feature interaction and dialogue context modeling.

\begin{table}[htb]
	\centering
	\caption{Experiment results on the IEMOCAP dataset. The best results are highlighted in bold.}
	 \setlength{\tabcolsep}{1.8mm}	
	\renewcommand{\arraystretch}{1.6}    
	\begin{tabular}{c|cccccc|c}
		\toprule
		Methods & Hap.  & Sad.  & Neu.  & Ang.  & Exc.  & Fru.  & Weighted F1 \\
		\midrule
		BC-LSTM \cite{Poria2017ContextDependentSA}  & 34.43 & 60.87 & 51.81 & 56.73 & 57.95 & 58.92 & 54.95 \\
		DialogueRNN \cite{Majumder2018DialogueRNNAA}  & 33.18 & 78.8  & 59.21 & 65.28 & 71.86 & 58.91 & 62.75 \\
		DialogueGCN \cite{Ghosal2019DialogueGCNAG}  & 51.87 & 76.76 & 56.76 & 62.26 & 72.71 & 58.04 & 63.16 \\
		IterativeERC \cite{Lu2020AnIE}  & 53.17 & 77.19 & 61.31 & 61.45 & 69.23 & 60.92 & 64.37 \\
		MMGCN \cite{Hu2021MMGCNMF}  & 42.34 & 78.67 & 61.73 & 69    & 74.33 & 62.32 & 66.22 \\
		MVN \cite{Ma2021AMN}  & 55.75 & 73.3  & 61.88 & 65.96 & 69.5  & 64.21 & 65.44 \\
		MM-DFN \cite{Hu2022MMDFNMD}  & 42.22 & 78.98 & 66.42 & 69.77 & 75.56 & 66.33 & 68.18 \\
		GA2MIF \cite{Li2022GA2MIFGA} & 46.15 & \textbf{84.5}  & 68.38 & 70.29 & 75.99 & 66.49 & 70 \\
		CRRGM \cite{Chen2024CRGMRAC}  & 52.4  & 76.8  & 70.1  & 65.4  & 77.3  & 63.6  & 70.4 \\
		TelME \cite{Yun2024TelMETM}  & -     & -     & -     & -     & -     & -     & 70.4 \\
		UniMSE \cite{Hu2022UniMSETU} & -     & -     & -     & -     & -     & -     & 70.66 \\
		MALN \cite{Ren2023MALNMA}  & 55.5  & 81.8  & 64.1  & 69.1  & \textbf{78}    & 71.4  & 70.8 \\
       \midrule
		Ours & \textbf{68.22} & 82.41 & \textbf{76.6}  & \textbf{77.35} & 67.01 & \textbf{71.9}  & \textbf{74.22} \\
		\bottomrule
	\end{tabular}%
	\label{tab:addlabel}%
\end{table}%

In MELD's fear detection, the performance reaches 43.42\%, an improvement of 16.45\% compared to TelME (26.97\%). Meanwhile, the cross-dataset anger recognition performance (IEMOCAP: 77.35\% vs. MELD: 64.45\%) ranked first, indicating that the model can effectively integrate the acoustic features of speech intonation with the implicit emotional cues of textual semantics.

\begin{table}[t]
	\centering
	\caption{Experiment results on the MELD dataset. The best results are highlighted in bold.}
	\setlength{\tabcolsep}{0.9mm}	
	\renewcommand{\arraystretch}{1.6}    
	\begin{tabular}{c|ccccccc|c}
		\toprule
		Methods & Neu.  & Sur.  & Fea.  & Sad.  & Joy   & Dis.  & Ang.  & Weighted F1 \\
		\midrule
		BC-LSTM \cite{Poria2017ContextDependentSA}    & 73.8  & 47.7  & 5.4  & 25.1  & 51.3  & 5.2   & 38.4  & 55.9 \\
		DialogueRNN \cite{Majumder2018DialogueRNNAA}  & 76.23 & 49.59 & 0   & 26.33 & 54.55 & 0.81  & 46.76 & 58.73 \\
		DialogueGCN \cite{Ghosal2019DialogueGCNAG}  & 76.02 & 46.37 & 0.98  & 24.32 & 53.62 & 1.22  & 43.03 & 57.52 \\
		IterativeERC \cite{Lu2020AnIE}  & 77.52 & 53.65 & 3.31  & 23.62 & 56.63 & 19.38 & 48.88 & 60.72 \\
		MMGCN \cite{Hu2021MMGCNMF}  & -     & -     & -     & -     & -     & -     & -     & 58.65 \\
		MVN \cite{Ma2021AMN}  & 76.65 & 53.18 & 11.7  & 21.82 & 53.62 & 21.86 & 42.55 & 59.03 \\
		MM-DFN \cite{Hu2022MMDFNMD}  & 77.76 & 50.69 & -     & 22.93 & 54.78 & -     & 47.82 & 58.65 \\
		EmoCaps \cite{Li2022EmoCapsEC} & 77.12 & 63.19 & 3.03  & 42.52 & 57.5  & 7.69  & 57.54 & 64 \\
		GA2MIF \cite{Li2022GA2MIFGA} & 76.92 & 49.08 & -     & 27.18 & 51.87 & -     & 48.52 & 58.94 \\
		TelME \cite{Yun2024TelMETM} & 80.22 & 60.33 & 26.97 & 43.45 & 65.67 & 26.42 & 56.7  & 67.37 \\
		UniMSE \cite{Hu2022UniMSETU} & -     & -     & -     & -     & -     & -     & -     & 65.51 \\
		MALN \cite{Ren2023MALNMA}  & 82    & 58.6  & 21.2  & 43    & 64.3  & 17.6  & 52.4  & 66.9 \\
        \midrule 
		Ours & \textbf{85.16} & \textbf{74.9}  & \textbf{43.42} & \textbf{46.83} & \textbf{67.94} & \textbf{28.37} & \textbf{64.45} & \textbf{73.1} \\
		\bottomrule
	\end{tabular}%
	\label{tab:addlabel}%
\end{table}%

\subsection{Ablation Study}
In this study, we conducted systematic ablation experiments, with the results shown in Table 3 and Table 4. The experiments demonstrate that after removing the PSA module, the model's Weighted F1 score on the IEMOCAP dataset decreased by 2.53\%, and on the MELD dataset by 0.55\%, with a significant decline in performance on fine-grained emotional categories such as happiness, fear, and disgust. Specifically, in the MELD dataset, the F1 score for the fear category dropped by 3.0\%, and for the disgust category by 2.28\%, highlighting the critical role of the PSA module in temporal feature modeling.

Furthermore, when both the MCN-CL framework is removed simultaneously, the model performance degrades significantly. The Weighted F1 score on the IEMOCAP dataset drops by 21.17\% (from 74.22\% to 53.05\%), and on the MELD dataset, it decreases by 2.74\% (from 73.10\% to 70.36\%). Particularly in extreme cases, such as the happy class on the IEMOCAP dataset, its F1 score drops sharply from 68.22\% to 2.41\%, while on the MELD dataset, the detection of fear and disgust categories completely fails (F1 score of 0). This result indicates that the multimodal cross-attention network can effectively capture cross-modal complementary information, and contrastive learning enhancement effectively alleviates the class imbalance problem.

\begin{table}[t]
	\centering
	\caption{Experiment results on the IEMOCAP dataset.}
	\setlength{\tabcolsep}{2.3mm}	
	\renewcommand{\arraystretch}{1.7}    
	\begin{tabular}{c|cccccc|c}
		\toprule
		Methods & Hap.  & Sad.  & Neu.  & Ang.  & Exc.  & Fru.  & Weighted F1 \\
		\midrule
		w/o PSA & 62.56 & 79.65 & 72.69 & 76.55 & 63.64 & 70.99 & 71.69 \\
	    w/o MCN-CL & 2.41  & 11.03 & 57.49 & 63.54 & 69.4  & 70.26 & 53.05 \\
		Ours  & 68.22 & 82.41 & 76.6  & 77.35 & 67.01 & 71.9  & 74.22 \\
		\bottomrule
	\end{tabular}%
	\label{tab:addlabel}%
\end{table}%

\begin{table}[t]
	\centering
	\caption{Experiment results on the MELD dataset.}
	\setlength{\tabcolsep}{1.6mm}	
	\renewcommand{\arraystretch}{1.7}    
	\begin{tabular}{c|ccccccc|c}
		\toprule
		Methods & Neu.  & Sur.  & Fea.  & Sad.  & Joy   & Dis.  & Ang.  & Weighted F1 \\
		\midrule
		w/o PSA & 84.79 & 74.28 & 40.45 & 45.73 & 67.69 & 26.09 & 64.04 & 72.55 \\
		w/o MCN-CL & 84.42 & 76.33 & 0     & 43.22 & 69.02 & 0     & 58.47 & 70.36 \\
		Ours  & 85.16 & 74.9  & 43.42 & 46.83 & 67.94 & 28.37 & 64.45 & 73.1 \\
		\bottomrule
	\end{tabular}%
	\label{tab:addlabel}%
\end{table}%

\section{Conclusions}
In this paper, we propose a novel method for multimodal emotion recognition using visual pyramid squeeze attention with contrastive cross-attention. The designed Pyramid Squeeze Attention module significantly enhances the ability to capture temporal features of facial action units, such as eyelid micro-tremors and lip muscle tension fluctuations, through multi-scale one-dimensional temporal convolution operations and channel attention interaction. Additionally, the multimodal cross-attention network and contrastive learning enhancement network, with a triple-query architecture and hard negative sample mining strategy, effectively resolve the acoustic-semantic coupling problem in highly confused categories, achieving an F1 score of 43.42\% for sparse categories (e.g., fear) on the MELD dataset. Experiment results on the IEMOCAP and MELD datasets demonstrate that our proposed method achieves Weighted F1 scores of 74.22\% and 73.10\%, respectively. In the future, we will advance multimodal learning to address high-frequency class bias, cross-modal noise decoupling, and modeling of extremely sparse samples.

\section*{Acknowledges}
This work was supported in part by the National Natural Science Foundation of China (Grant No. 12574510), the Beijing Natural Science Foundation (No. L257021), the Key Project of Anhui Provincial University Scientific Research Plan (No. 2024AH050018), and the Key Project in Natural Sciences of Anhui University of Finance and Economics (No. ACKYB23016).

\end{document}